\documentclass{article}

\usepackage{microtype}
\usepackage{graphicx}
\usepackage{subcaption}
\usepackage{booktabs}
\usepackage{hyperref}

\usepackage[accepted]{icml2026}

\usepackage{amsmath,amssymb,amsfonts}
\usepackage{xcolor}
\usepackage{multirow}
\usepackage[capitalize,noabbrev]{cleveref}

\icmltitlerunning{Prompt Optimization Is a Coin Flip}

\begin{document}

\twocolumn[
\icmltitle{Prompt Optimization Is a Coin Flip:\\Diagnosing When It Helps in Compound AI Systems}

\begin{icmlauthorlist}
\icmlauthor{Xing Zhang}{aws}
\icmlauthor{Guanghui Wang}{aws}
\icmlauthor{Yanwei Cui}{aws}
\icmlauthor{Wei Qiu}{hsbc}
\icmlauthor{Ziyuan Li}{hsbc}
\icmlauthor{Bing Zhu}{hsbc}
\icmlauthor{Peiyang He}{aws}
\end{icmlauthorlist}

\icmlaffiliation{aws}{AWS Generative AI Innovation Center}
\icmlaffiliation{hsbc}{HSBC Holdings Plc., HSBC Technology Center, China}

\icmlcorrespondingauthor{Peiyang He}{peiyan@amazon.com}

\icmlkeywords{compound AI systems, prompt optimization, multi-agent, ANOVA, interaction effects, model specificity, evaluation methodology}

\vskip 0.3in
]

\printAffiliationsAndNotice{}

\begin{abstract}

Prompt optimization in compound AI systems is statistically indistinguishable from a coin flip: across 72 optimization runs on Claude Haiku~4.5 (6 methods $\times$ 4 tasks $\times$ 3 repeats), 49\% score \emph{below} zero-shot; on Amazon Nova Lite, the failure rate is even higher.
Yet on one task, \emph{all} six methods improve over zero-shot by up to $+6.8$ points.
What distinguishes success from failure?
We investigate with 18,000 grid evaluations and 144 optimization runs, testing two assumptions behind end-to-end optimization tools like TextGrad and DSPy, in the order they must be answered: (A)~agent prompts interact, requiring joint rather than independent optimization, and (B)~individual prompts are worth optimizing at all.
Interaction effects are never significant ($p > 0.52$, all $F < 1.0$), and optimization helps only when the task has \emph{exploitable output structure}: a format the model can produce but does not default to.
We further give a mechanistic account: instruction-tuning compresses input phrasing into a narrow output distribution, eliminating the very phrasing-sensitivity that joint optimization assumes.
We provide a two-stage diagnostic: an \$80 ANOVA pre-test for agent coupling, and a 10-minute headroom test that predicts whether optimization is worthwhile, turning a coin flip into an informed decision.

\end{abstract}

\section{Introduction}
\label{sec:intro}

Compound AI systems (pipelines of multiple LLM calls where each agent handles a specialized subtask) have become the dominant architecture for complex applications \citep{chase2022langchain, wu2023autogen}.
A natural question arises: \emph{how should we optimize the prompts in these systems?}

Recent work strongly favors end-to-end joint optimization.
TextGrad \citep{yuksekgonul2024textgrad} propagates textual gradients through multi-component systems.
DSPy \citep{khattab2023dspy} compiles LLM programs with end-to-end optimization.
GPTSwarm \citep{zhuge2024gptswarm} treats agent graphs as optimizable structures.
These methods implicitly rely on two assumptions:

\begin{itemize}
    \item \textbf{Assumption A (coupling)}: Agent prompts interact, so the optimal prompt for one agent depends on the prompt of another, requiring joint rather than independent optimization.
    \item \textbf{Assumption B (worth optimizing)}: Individual agent prompts are worth optimizing, in that changing a prompt meaningfully affects system output, even at realistic training budgets.
\end{itemize}

If Assumption~A fails, independent per-agent optimization suffices.
If Assumption~B also fails, even per-agent optimization is unnecessary.
Surprisingly, neither assumption has been empirically tested.

We provide controlled measurements of both, in the order they must be answered, and identify the conditions that separate the 49\% failure rate from the cases where optimization delivers real gains.
\textbf{Study~1} (\S\ref{sec:coupling}) tests Assumption~A via exhaustive grid evaluation of two-agent pipelines with ANOVA decomposition: \emph{do we even need joint optimization?}
\textbf{Study~2} (\S\ref{sec:optimization}) tests Assumption~B by benchmarking six optimization methods against zero-shot baselines under equal compute budgets: \emph{given that joint optimization is unnecessary, does individual optimization help?} We then analyze \emph{why} it works when it does (\S\ref{sec:when_works}).

\paragraph{Key findings.}
\begin{enumerate}
    \item \textbf{Agents don't interact.} The $A \times B$ interaction term is non-significant in all six model$\times$task conditions ($p > 0.52$, all $F < 1.0$); joint optimization is unnecessary (\S\ref{sec:coupling}). \S\ref{sec:discussion} explains why this is mechanistically expected from instruction-tuning.
    \item \textbf{Optimization helps only when exploitable structure exists.} The sole task where all methods succeed requires structured rubrics and JSON formatting, a format the model \emph{can} produce but does not default to. The other three tasks lack this gap (\S\ref{sec:when_works}).
    \item \textbf{Model specificity dominates.} Which agents matter, which tasks benefit, and which methods work all change with the executor model (\S\ref{sec:model_specificity}).
    \item \textbf{A 10-minute test predicts optimization-worthiness.} Generate 10--20 candidate prompts; when the best candidate gains $<$2 pts over zero-shot, the landscape is flat and none of the six methods we tested reliably help. The 2-pt threshold is calibrated to our setup; the qualitative principle (flat landscape implies no useful headroom) is general (\S\ref{sec:framework}).
\end{enumerate}

\begin{figure*}[t]
\centering
\includegraphics[width=\textwidth]{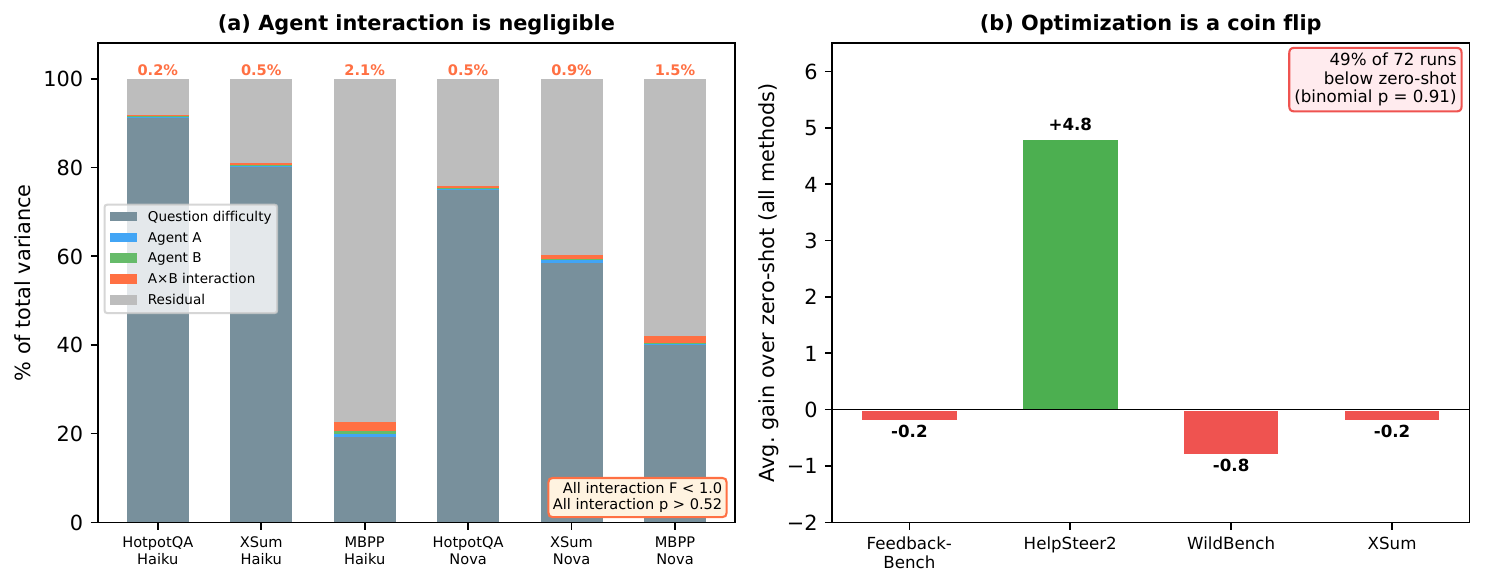}
\caption{\textbf{Overview.} (a)~Variance decomposition across 6 model$\times$task conditions (Study~1, \S\ref{sec:coupling}). Interaction (orange) accounts for only 0.18--2.15\% of total variance; all $F < 1.0$. (b)~Average gain of optimization methods over zero-shot (Study~2, \S\ref{sec:optimization}). Only HelpSteer2 shows positive gains; it is the sole task with exploitable output structure (\S\ref{sec:when_works}). On the other 3 tasks, average gains are negative; 49\% of 72 Haiku runs score below zero-shot (binomial $p = 0.91$).}
\label{fig:hero}
\end{figure*}

Figure~\ref{fig:hero} summarizes both studies; we distill the findings into a two-stage diagnostic framework: an \$80 ANOVA pre-test for coupling, and a headroom test for optimization-worthiness (\S\ref{sec:framework}).

\emph{Important caveat:} our findings do not imply that compound AI systems are ineffective, nor that prompt optimization is universally useless. They show that, in our setup, we cannot reject the hypothesis that optimization performs no better than random; the practical implication is that optimization must be targeted at the right conditions, and that expensive joint optimization is unnecessary.
More broadly, our ANOVA-based methodology offers a principled evaluation protocol for measuring whether capabilities compose in multi-agent systems: a structured alternative to ad hoc benchmarking.

\section{Related Work}
\label{sec:related}

\paragraph{Compound AI optimization.}
TextGrad~\citep{yuksekgonul2024textgrad}, DSPy~\citep{khattab2023dspy}, and GPTSwarm~\citep{zhuge2024gptswarm} assume components interact enough to justify joint optimization; we provide the first empirical test of this assumption.
Helix~\citep{zhu2026helix} co-evolves prompts and queries, a joint optimization approach whose premise our Study~1 calls into question.
A recent survey~\citep{yue2026workflow} systematizes agentic workflow optimization into static and dynamic paradigms but does not empirically test whether the inter-agent coupling that motivates these methods exists.
Single-call prompt optimization (APE~\citep{zhou2023ape}, OPRO~\citep{yang2024opro}, PromptBreeder~\citep{fernando2024promptbreeder}, ProTeGi~\citep{pryzant2023automatic}, EvoPrompt~\citep{guo2023connecting}) does not address inter-agent dependencies.
Multi-agent architectures~\citep{wu2023autogen, wang2024mixture, hu2024adas} and iterative self-refinement~\citep{shinn2023reflexion} propose increasingly sophisticated designs; our work measures whether the inter-agent coupling these architectures create is strong enough to require joint optimization.

\paragraph{Prompt optimization benchmarks.}
Prior comparisons evaluate methods on different tasks with different budgets.
To our knowledge, no existing comparison uses strictly equal compute budgets across six methods with two executor models and tests whether gains are model-specific.

\section{Study 1: Measuring Agent Coupling via ANOVA}
\label{sec:coupling}

\subsection{Method}

We study two-agent pipelines ($\text{Agent A} \rightarrow \text{Agent B}$) where Agent~A processes the input and Agent~B produces the final output using Agent~A's response.
For each of three tasks, we generate $K{=}10$ diverse candidate system prompts per agent (varying strategy, tone, and structure) and exhaustively evaluate all $10 \times 10 = 100$ prompt combinations on $n{=}30$ benchmark samples, yielding a score tensor $Y_{ijk}$.

\paragraph{Tasks.}
We deliberately select tasks with varying \emph{a priori} expected coupling:
\textbf{HotpotQA} \citep{yang2024hotpotqa} (multi-hop QA; expected: tight coupling),
\textbf{MBPP} \citep{austin2021mbpp} (code generation; expected: medium),
\textbf{XSum} \citep{narayan2018xsum} (summarization; expected: loose).

\paragraph{Models.}
Claude Haiku~4.5 (mid-tier) and Amazon Nova Lite (budget-tier) as executors; Claude Sonnet~4.6 as judge.

\paragraph{Analysis.}
Two-way ANOVA with question blocking decomposes total variance into five sources: question difficulty, Agent~A main effect, Agent~B main effect, $A{\times}B$ interaction, and residual.
The interaction $F$-test directly answers: \emph{does the optimal prompt for one agent depend on the other?}

\subsection{Results: Interaction Is Never Significant}

\begin{table}[t]
\caption{Variance decomposition (\% of total sum of squares) for the two-agent pipeline grid ($10{\times}10$ Agent~A $\times$ Agent~B prompts, $n{=}30$ questions per cell). Columns: \textbf{Q}~=~question difficulty (block effect); \textbf{A}~=~Agent~A main effect; \textbf{B}~=~Agent~B main effect; \textbf{A$\times$B}~=~interaction (the term joint optimization assumes is large); \textbf{Err}~=~residual. The interaction term is never statistically significant and never exceeds the residual mean square; all $F < 1.0$.}
\label{tab:anova}
\vskip 0.1in
\centering
\footnotesize
\newcommand{\sig}[2]{\makebox[3.2em][l]{#1\textsuperscript{\scriptsize #2}}}
\newcommand{\nosig}[1]{\makebox[3.2em][l]{#1}}
\setlength{\tabcolsep}{3pt}
\begin{tabular}{@{}ll r cc r r@{}}
\toprule
\textbf{Model} & \textbf{Task} & \textbf{Q} & \textbf{A} & \textbf{B} & \textbf{A$\times$B} & \textbf{Err} \\
\midrule
\multirow{3}{*}{Haiku}
& HotpotQA & 91.3 & \sig{0.05}{*} & \sig{0.37}{***} & 0.18 & 8.1 \\
& XSum     & 80.3 & \nosig{0.09} & \nosig{0.09} & 0.49 & 19.0 \\
& MBPP     & 19.3 & \sig{0.60}{**} & \sig{0.59}{**} & 2.15 & 77.4 \\
\midrule
\multirow{3}{*}{Nova}
& HotpotQA & 75.1 & \nosig{0.12} & \nosig{0.08} & 0.51 & 24.2 \\
& XSum     & 58.4 & \sig{0.77}{***} & \nosig{0.22} & 0.87 & 39.7 \\
& MBPP     & 39.9 & \sig{0.45}{**} & \nosig{0.16} & 1.50 & 58.0 \\
\bottomrule
\end{tabular}
\vskip 0.05in
\raggedright\footnotesize \textsuperscript{*}$p{<}0.05$, \textsuperscript{**}$p{<}0.01$, \textsuperscript{***}$p{<}0.001$. Interaction $p > 0.52$ in all conditions; all $F < 1.0$.
\end{table}

Table~\ref{tab:anova} presents the central finding: the $A \times B$ interaction term is non-significant in every condition, accounting for 0.18--2.15\% of total variance with all $F < 1.0$.
The interaction mean square does not exceed the residual mean square in any condition.

\paragraph{Expert predictions were wrong.}
HotpotQA was expected to be tightly coupled, since multi-hop reasoning seemingly requires Agent~B to build on Agent~A's specific decomposition.
It shows the \emph{smallest} interaction (0.18\% on Haiku).
Even practitioners cannot predict coupling, making empirical measurement essential.
We return to why the absence of coupling is mechanistically expected (rather than an accident of our task selection) in \S\ref{sec:discussion}.

\paragraph{Question difficulty dominates.}
Question difficulty explains 19--91\% of total variance, far more than any prompt effect.
Evaluation noise from question sampling may overwhelm real optimization signals.

\begin{figure*}[t]
\centering
\includegraphics[width=\textwidth]{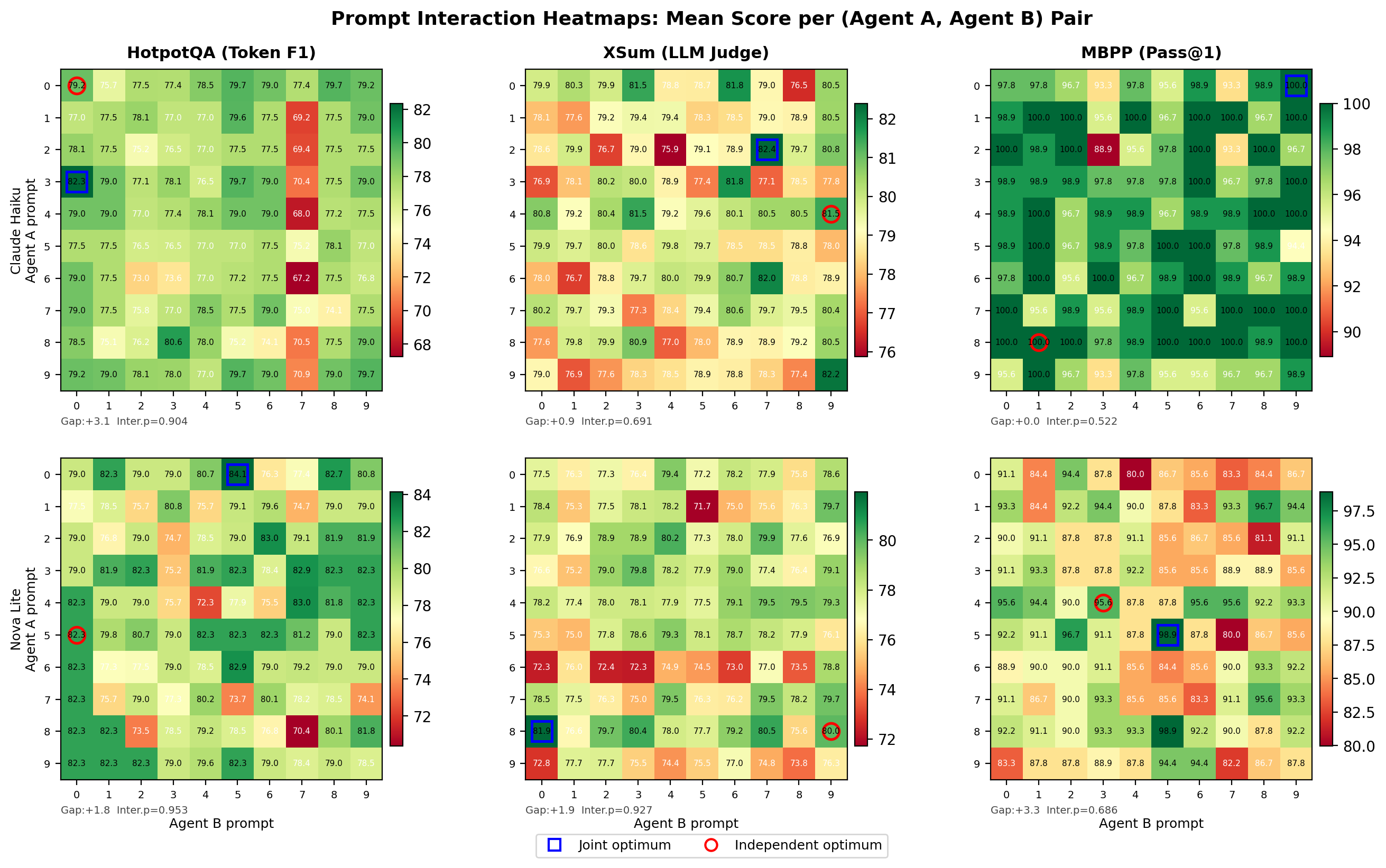}
\caption{Score matrices for all 6 model$\times$task conditions ($10 \times 10$ Agent~A $\times$ Agent~B prompt grid; cell color = mean score over $n{=}30$ questions). \textbf{Row/column banding dominates}, with no off-diagonal synergy: the visual signature of near-zero $A{\times}B$ interaction (Table~\ref{tab:anova}). Blue square = joint optimum; red circle = independent optimum (best row $\times$ best column). The two are adjacent or identical in every condition; gap 0.0--3.3 pts.}
\label{fig:heatmaps}
\end{figure*}

\paragraph{Visual evidence: additive structure.}
Figure~\ref{fig:heatmaps} shows the score matrices for all six model$\times$task conditions.
The dominant pattern is \emph{row and column banding}: good Agent~A prompts produce consistently good rows, and good Agent~B prompts produce consistently good columns.
There is no off-diagonal structure: no prompt pair works synergistically beyond the sum of its parts.
This additive pattern is the visual signature of near-zero interaction.
The joint optimum (blue square) and independent optimum (red circle) are adjacent or identical in every condition.

\paragraph{Joint vs.\ independent optimum.}
The gap between the jointly-optimal prompt pair (best cell in the grid) and the independently-optimal pair (best row $\times$ best column) ranges from 0.0 to 3.3 points: an upper bound that real methods would not reach.
Budget-equalized simulations (1,000 trials) confirm that independent search matches joint search at all budget levels.

\paragraph{The interaction residual is structurally random.}
Beyond the $F$-test, we examine the residual landscape after subtracting additive row/column effects: the part that any joint optimizer would have to exploit.
Neighbor autocorrelation in this residual is near zero ($\rho \in [-0.12, +0.05]$ across the six conditions), indicating the surface is statistically indistinguishable from random noise rather than a smooth gradient.
This directly undermines the assumption underlying gradient-based joint optimization (e.g., TextGrad): there is no smooth signal to propagate.
We discuss the mechanistic explanation in \S\ref{sec:discussion}.

\section{Study 2: When Does Individual Optimization Help?}
\label{sec:optimization}

If joint optimization is unnecessary (Study~1), perhaps independent per-agent optimization suffices?
We test this on four single-agent tasks chosen for diversity in output format and evaluation metric.
Study~2 isolates single-agent optimization to test Assumption~B without confounding pipeline effects; XSum appears in both studies, providing a direct bridge.

\subsection{Method}

We compare six methods (APE \citep{zhou2023ape}, OPRO \citep{yang2024opro}, EvoPrompt \citep{guo2023connecting}, PromptBreeder \citep{fernando2024promptbreeder}, DSPy-style bootstrap \citep{khattab2023dspy}, and PROSE\footnote{PROSE (PRompt Optimization via Structured Evolution) is our own method that adds risk-aware selection to evolutionary prompt search. We include it to test whether explicit risk-aware selection helps; it does not (Appendix~\ref{app:prose}), reinforcing our main finding.}) against zero-shot and manual baselines.
All methods evaluate ${\sim}$100 candidate prompts (equal compute budget).
Four tasks: Feedback-Bench \citep{kim2024prometheus}, HelpSteer2 \citep{wang2024helpsteer2}, WildBench \citep{lin2024wildbench}, and XSum \citep{narayan2018xsum}.
Training: 20 questions; test: 100 held-out questions; 3 repeats per condition.
Two executor models: Claude Haiku~4.5 and Amazon Nova Lite.

\subsection{Results: Optimization Is a Coin Flip}

\begin{table}[t]
\centering
\caption{Held-out test scores on Claude Haiku~4.5 (mean over 3 repeats; 100 test questions per task; LLM-judge scale 0--100). Tasks: \textbf{FB}~=~Feedback-Bench, \textbf{HS2}~=~HelpSteer2, \textbf{WB}~=~WildBench, \textbf{XSum}~=~XSum summarization. \textbf{Bold} = best method per task. HS2 is the only task where every optimizer beats zero-shot; on the other three, average gain is negative or near zero.}
\label{tab:optimization}
\small
\begin{tabular}{@{}lcccc@{}}
\toprule
Method & FB & HS2 & WB & XSum \\
\midrule
Zero-Shot & 82.4 & 68.0 & 68.9 & 76.0 \\
\midrule
APE           & 82.3 & 69.3 & 68.0 & \textbf{76.6} \\
OPRO          & 81.4 & 73.8 & 69.0 & 74.7 \\
EvoPrompt     & 82.0 & \textbf{74.8} & 68.3 & 75.6 \\
PromptBreeder & \textbf{83.5} & 74.6 & 68.5 & 76.0 \\
DSPy-style    & 81.9 & 69.8 & 65.1 & 76.2 \\
PROSE         & 82.1 & 74.4 & \textbf{69.6} & 75.9 \\
\bottomrule
\end{tabular}
\end{table}

Table~\ref{tab:optimization} shows the results.
On Haiku, 49\% of 72 runs (6 methods $\times$ 4 tasks $\times$ 3 repeats) score \emph{below} zero-shot.
We cannot reject the null hypothesis that gain over zero-shot is symmetric around zero (binomial $p = 0.91$): in our setup, optimization is statistically indistinguishable from random selection at the binomial level.
On Nova Lite, the picture is worse: 14 of 24 method$\times$task means fall below zero-shot (Table~\ref{tab:nova} in Appendix~\ref{app:nova}).
On three of four tasks, the average gain across all methods is \emph{negative}: $-0.20$ (FB), $-0.82$ (WB), $-0.17$ (XSum).

\paragraph{Why optimization usually fails.}
Two factors compound: with only 20 training questions, per-candidate scores are too noisy for reliable selection, and iterative methods overfit (train-test gaps up to $+5.6$ pts; APE, being non-iterative, shows none).
This aligns with broader evidence: \citet{nie2026igo} report that only 9\% of surveyed agents use any automated optimization, attributing low adoption to hidden design choices that compound the noise problem we observe.

\subsection{When Does Optimization Work?}
\label{sec:when_works}

The aggregate coin-flip statistic conceals a striking exception: on HelpSteer2, \emph{all} six methods beat zero-shot on Haiku (best $\Delta = +6.8$ pts from EvoPrompt), and all methods reaching $\geq$74 pts independently discover the same structure.
Understanding \emph{why} this task succeeds is more actionable than the aggregate failure rate.
We caution that the ``exploitable structure'' explanation we develop next is a hypothesis derived from a single positive case (HelpSteer2): it explains the present results but has not been validated on held-out tasks. Our 10-minute headroom test (\S\ref{sec:framework}) is the operational form of this hypothesis, applied prospectively rather than retrospectively.

\paragraph{The ``can but doesn't'' pattern.}
HelpSteer2 requires structured rubric-based evaluation with JSON-formatted output.
Haiku \emph{can} produce this format (when prompted correctly, scores jump from 68.0 to 74.8), but its zero-shot default is unstructured prose.
The optimization landscape has a clear feature to exploit: a specific output format the model knows but does not default to.
We hypothesize this pattern generalizes to tasks requiring specific output schemas (JSON, XML), domain-specific formatting conventions, or structured reasoning templates: any setting where the model has latent capability that a well-chosen prompt can unlock.

\paragraph{Why the other tasks fail.}
Feedback-Bench, WildBench, and XSum accept free-form natural language output.
The model's zero-shot behavior is already near-optimal for the format these tasks require; there is no latent capability gap for optimization to unlock.
Gains over zero-shot confirm this: HelpSteer2's best method gains $+6.8$ pts, while the best methods on Feedback-Bench, WildBench, and XSum gain only $+1.1$, $+0.7$, and $+0.6$ pts respectively, within the noise floor of 20-question evaluation.

\paragraph{Diagnostic: the headroom test.}
This analysis suggests a fast, cheap diagnostic for optimization-worthiness:
generate 10--20 candidate prompts and compare their best score to zero-shot on 20~held-out questions.
In our data, a $>$2~pt gain cleanly separates the one successful task from the three failures, while a $<$2~pt gain reliably indicates a flat landscape where no method we tested helps.
The test takes ${\sim}$10 minutes and ${\sim}$\$5.
We stress that the specific 2-point threshold is calibrated to our setup (two mid-tier executor models, 20-question training sets, six optimizers, and an LLM-judge metric on a 0--100 scale), and practitioners should re-calibrate against their own zero-shot noise floor before treating it as a hard cutoff. The qualitative claim that we expect to generalize is the underlying signal: \emph{on a flat landscape, modest random search already exposes the ceiling}.

\section{The Dominant Factor Is Model, Not Prompt}
\label{sec:model_specificity}

Both studies reveal the same meta-finding: \emph{everything changes when the model changes}.

\paragraph{Study 1: which agent matters flips.}
On Haiku, Agent~B (synthesizer) dominates HotpotQA ($p < 0.001$); on Nova, neither agent is significant.
On XSum, Agent~A (extractor) matters on Nova ($p < 0.001$) but not Haiku.
The bottleneck agent depends on the model, not the task architecture.

\paragraph{Study 2: which task is optimizable reverses.}
On HelpSteer2, all 6 methods beat zero-shot on Haiku (Table~\ref{tab:optimization}, best $\Delta = +6.8$); on Nova Lite, only 1 of 6 does (Table~\ref{tab:nova}, best $\Delta = +2.1$).
The same task moves from highly optimizable to flat by changing the executor model alone.
Meanwhile, Feedback-Bench goes from 1/6 on Haiku to 4/6 on Nova: a complete reversal.

\paragraph{Implication.}
Neither coupling structure nor optimization headroom can be determined \emph{a priori}; both are empirical properties of the specific model--task combination.
This is precisely why a cheap diagnostic is valuable: rather than assuming optimization will help (or won't), measure it for your specific setting.

\section{Practitioner Framework}
\label{sec:framework}

\begin{figure}[t]
\centering
\includegraphics[width=\columnwidth]{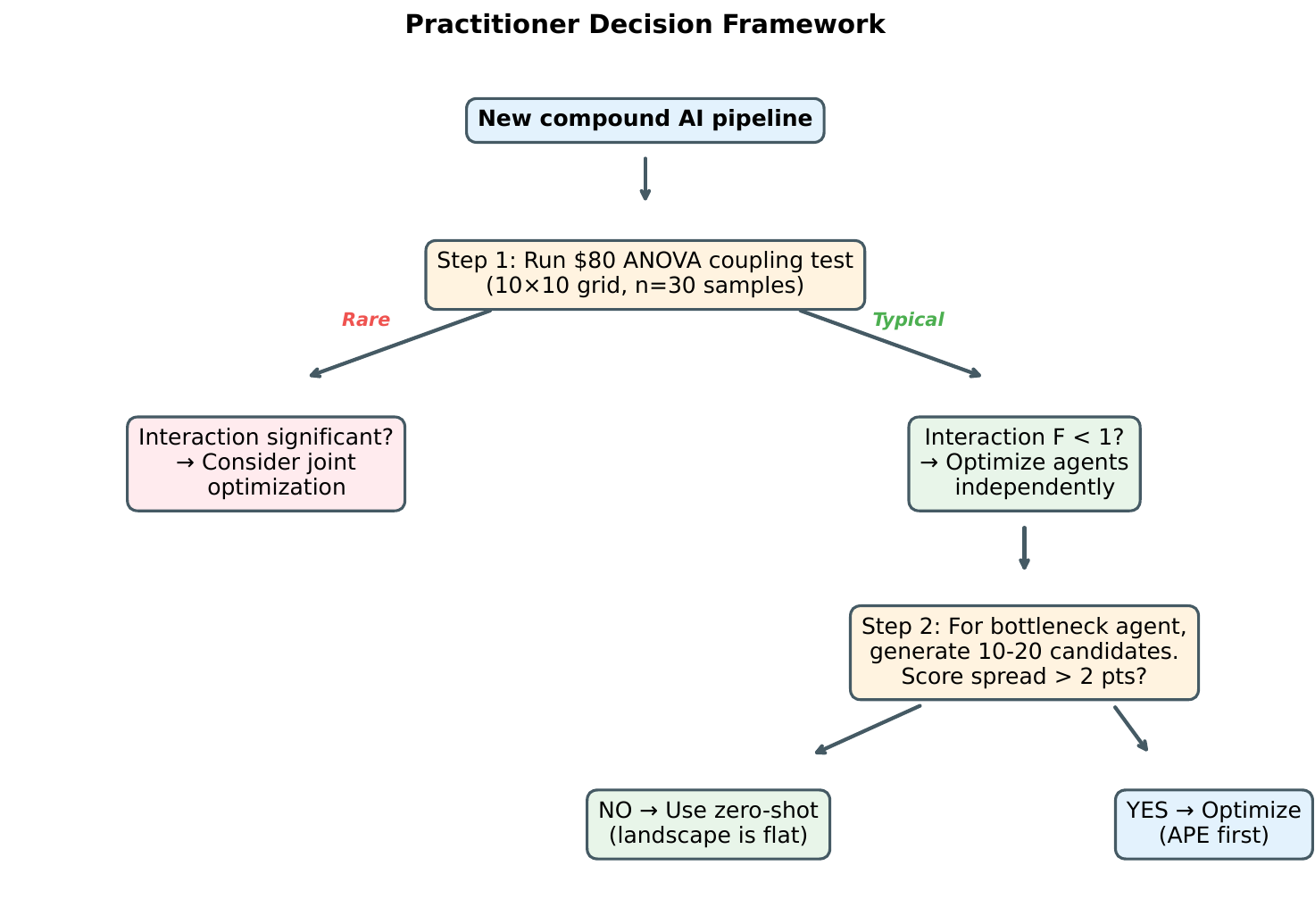}
\caption{Practitioner decision framework. \textbf{Stage~1} (\$80, 1 day): ANOVA grid (Table~\ref{tab:anova}); if interaction $F < 1$, agents are decoupled. \textbf{Stage~2} (\$5, 10 min): generate 10--20 candidates; if best gain $>$2~pts over zero-shot, optimize with APE-style generate-and-rank, else use zero-shot. Cost comparison vs.\ DSPy/TextGrad: Table~\ref{tab:cost}.}
\label{fig:framework}
\end{figure}

We distill our findings into a two-stage diagnostic protocol (Figure~\ref{fig:framework}).

\textbf{Prerequisite: Model selection.} Choose the right model first; model selection dominates all prompt-level optimization in our data. Always re-run the following stages after model updates.

\textbf{Stage 1: Coupling test} (\$80, 1 day). Run the ANOVA grid ($10{\times}10$ prompts, $n{=}30$ samples) to measure agent interaction. If $F < 1$, agents are decoupled; optimize independently. Use main effects to identify the bottleneck agent.

\textbf{Stage 2: Headroom test} (\$5, 10 min). For the bottleneck agent, generate 10--20 candidate prompts. If the best candidate gains $>$2 pts over zero-shot, look for the ``can but doesn't'' pattern (\S\ref{sec:when_works}) and optimize with APE-style generate-and-rank (no overfitting risk). If the gain is $<$2 pts, the landscape is flat; use zero-shot.

\paragraph{Cost-benefit comparison.}
Table~\ref{tab:cost} contextualizes the framework.
The two-stage diagnostic costs ${\sim}$\$85 total and takes 1--2 days; if it reveals no coupling and no headroom (as in 3 of our 4 tasks), teams avoid committing \$1{,}000--10{,}000+ to optimization methods that assume otherwise.

\begin{table*}[t]
\centering
\caption{Cost comparison of optimization approaches. The Stage~1 and Stage~2 figures are measured on our experiments (mid-tier models, 3 tasks, $n{=}30$); the DSPy and TextGrad rows are \emph{order-of-magnitude} estimates extrapolated from each tool's reported per-iteration call counts and current Bedrock/API list prices, and should be read as relative magnitudes rather than precise quotes. Actual cost will vary with task length, iteration count, and judge-model choice.}
\label{tab:cost}
\small
\begin{tabular}{@{}lrll@{}}
\toprule
\textbf{Approach} & \textbf{Cost} & \textbf{What you learn} & \textbf{When to use} \\
\midrule
Stage 1: ANOVA coupling test & ${\sim}$\$80   & Do agents interact? & Always; rules out joint optimization if $F < 1$ \\
Stage 2: Headroom test & ${\sim}$\$5  & Is optimization worthwhile? & After Stage 1; checks if landscape is flat \\
\quad + APE generate-and-rank & ${\sim}$\$20  & Best single-agent prompt & Optional extension if Stage 2 finds headroom \\
\midrule
DSPy compilation    & \$1--5K  & Compiled pipeline & Only if coupling test shows significant interaction \\
TextGrad end-to-end & \$5--10K & Joint-optimized prompts & Only if coupling test shows significant interaction \\
\bottomrule
\end{tabular}
\end{table*}

\section{Discussion}
\label{sec:discussion}

\paragraph{Why are interactions weak?}
Instruction-tuning and RLHF train models to produce consistent outputs across diverse input phrasings, effectively compressing a wide range of input styles into a narrow output distribution.
In a two-agent pipeline, this means Agent~B's output variance is dominated by the \emph{semantic content} of Agent~A's response (which is determined by the question), not by Agent~A's \emph{stylistic variation} (which is what prompt changes affect).
The pipeline behaves as a composition of independently-robust functions: coupling requires agents to depend on each other's phrasing, but instruction-tuning specifically eliminates phrasing sensitivity.

\paragraph{The interaction landscape is noise (recap).}
\S\ref{sec:coupling} reported neighbor autocorrelation $\rho \in [-0.12, +0.05]$ on the residual landscape.
Combined with the instruction-tuning argument above, this gives both a statistical and a mechanistic reason to expect the additive structure: the residual is random because the underlying agent functions, post instruction-tuning, are independently robust to phrasing.

\paragraph{Realistic practitioner budgets.}
One might argue that 20 training questions is too few for optimization, and that this stacks the deck against iterative methods which spend most of their budget chasing noise.
We agree this is a real constraint, but it is the constraint most practitioners face.
If prompt optimization requires hundreds of labeled examples to work reliably, this is itself an important negative result: the methods are not practical under the budgets they implicitly target.
We expect that with substantially larger training sets the iterative methods would close some of the gap to APE, and we leave a budget-sweep ablation to future work; what our results do establish is that under realistic-budget conditions, the gap to zero-shot is what dominates, and that gap is structural (the ``can but doesn't'' pattern), not a function of training-set size.

\paragraph{Model specificity has growing consequences.}
In a landscape where frontier models update quarterly, our finding that optimization effects are model-specific is arguably more consequential than the independence result itself: any prompt optimization strategy has a shelf life shorter than the model release cycle.
Teams that invest \$10K in TextGrad-optimized prompts for Model~X face re-optimization costs when Model~X+1 arrives, and our data shows that which agents matter, which tasks benefit, and which methods work may all reverse.
This does not mean tools like DSPy or TextGrad are methodologically wrong; rather, the base models are rapidly absorbing the very tricks these tools were designed to discover.
Scaffold techniques that once required careful prompt engineering (chain-of-thought decomposition, structured output formatting, ReAct-style tool use) are increasingly built into model capabilities through RL training, shrinking the optimization headroom that external tools can exploit.
Our diagnostic framework is cheap enough to re-run after every model update, tracking this shrinking headroom over time.

\paragraph{When might independence break?}
Our study covers two-agent feed-forward pipelines with mid-tier models.
We predict coupling may emerge when: (1)~agents share mutable state (e.g., a common database or scratchpad); (2)~Agent~B's input format depends on Agent~A's output schema; (3)~pipelines have feedback loops (Agent~B's output feeds back to Agent~A); (4)~deeper pipelines (3+ agents) accumulate interaction opportunities; or (5)~agents communicate via structured data (JSON schemas, code) rather than natural language, where format sensitivity may amplify coupling.
Architectures combining several of these properties, such as VMAO's verification-driven replanning over shared DAG state~\citep{zhang2026vmao}, are natural candidates for coupling measurement.
Critically, our \$80 ANOVA pre-test generalizes to any of these architectures: practitioners should run it on \emph{their} pipeline rather than rely on our results or their intuitions, which we have shown to be unreliable.

\paragraph{Toward structured evaluation of compound AI.}
Our ANOVA decomposition produces falsifiable predictions that can be tested on any architecture: if a pipeline's interaction $F > 1$, joint optimization is warranted; if $F < 1$, it is not.
This contrasts with typical compound AI evaluation, which reports aggregate task scores without decomposing \emph{why} a pipeline succeeds or fails.
We hope the methodology generalizes beyond prompt optimization to any setting where practitioners need to measure whether components of a compound system compose or operate independently.

\paragraph{Limitations.}
Our grid uses $K{=}10$ whole-prompt substitutions; finer-grained edits (single-constraint flips, structured component substitutions) could in principle expose interaction patterns that whole-prompt swaps mask, and we view the present result as evidence against \emph{coarse} prompt-level coupling specifically.
Both executor models (Claude Haiku~4.5, Amazon Nova Lite) are mid-tier; a frontier-tier executor would strengthen the model-specificity finding by triangulating the trend.
Study~2 uses 20 training questions, reflecting practitioner reality but potentially limiting iterative methods, and Study~2's optimization tasks (FB/HS2/WB/XSum) overlap with Study~1 only on XSum; running optimization on the HotpotQA and MBPP pipelines would directly test whether the single-agent vs.\ two-agent design itself shifts optimization-worthiness, which we leave to future work.
Both studies use a single two-agent, feed-forward architecture; the hypotheses enumerated above about three-agent pipelines, shared scratchpads, feedback loops, and structured-data communication are plausible but untested here.

\section{Conclusion}
\label{sec:conclusion}

Prompt optimization in compound AI systems is not universally futile, but it is highly conditional.
We tested the two core assumptions underlying end-to-end optimization and found that agents do not interact (all $F < 1.0$ across 18,000 evaluations) and that individual optimization helps only when the task has exploitable output structure: the ``can but doesn't'' pattern.
On the one task with this pattern, all six methods succeed; aggregating across all four tasks, 49\% of runs score below zero-shot.

The key contribution is not the negative result, but the evaluation methodology: ANOVA variance decomposition provides a principled, falsifiable test for compositional behavior in multi-agent systems, and the headroom test diagnoses optimization-worthiness before committing resources.
Together, they turn a coin flip into an informed decision:

\paragraph{Recommendations.}
\begin{enumerate}
    \item \textbf{Test for coupling (\$80, 1 day).} Run the ANOVA grid on your pipeline. If $F < 1$ for the interaction term, optimize agents independently and do not invest in joint optimization.
    \item \textbf{Test for headroom (\$5, 10 min).} Generate 10--20 candidate prompts. If the best gains $<$2 pts over zero-shot, the landscape is flat; use zero-shot. If the gain is $>$2 pts, look for the ``can but doesn't'' pattern: a specific output format or structure the model knows but does not default to.
    \item \textbf{Re-test after every model update.} Which agents matter, which tasks benefit, and which methods work all change with the model. Budget optimization as recurring, not one-time.
\end{enumerate}

\section*{Impact Statement}

This paper presents work whose goal is to advance the field of compound AI system optimization.
Our diagnostic framework helps practitioners avoid wasted computation on ineffective prompt optimization, reducing both cost and environmental impact.
We find no ethical concerns specific to this work beyond those generally associated with advancing prompt engineering capabilities.

\newpage
\appendix

\section{ANOVA Formulas}
\label{app:anova}

Let $Y_{ijk}$ denote the score for Agent~A prompt $i$, Agent~B prompt $j$, and question $k$.
We subtract question means to obtain $\tilde{Y}_{ijk} = Y_{ijk} - \bar{Y}_{\cdot\cdot k}$, then decompose:
\begin{align}
\text{SS}_A &= nK_B \sum_i (\tilde{\bar{Y}}_{i\cdot\cdot} - \tilde{\bar{Y}}_{\cdot\cdot\cdot})^2 \\
\text{SS}_B &= nK_A \sum_j (\tilde{\bar{Y}}_{\cdot j\cdot} - \tilde{\bar{Y}}_{\cdot\cdot\cdot})^2 \\
\text{SS}_{A\times B} &= n\sum_{i,j} (\tilde{\bar{Y}}_{ij\cdot} - \tilde{\bar{Y}}_{i\cdot\cdot} - \tilde{\bar{Y}}_{\cdot j\cdot} + \tilde{\bar{Y}}_{\cdot\cdot\cdot})^2
\end{align}
The interaction has $(K_A{-}1)(K_B{-}1) = 81$ degrees of freedom, vs.\ 9 for each main effect.
This is why 2\% of total variance can be non-significant: $\text{MS}_{A\times B} = 2\%/81$ per df, while $\text{MS}_A = 0.6\%/9$ per df yields a larger $F$.

\section{Nova Lite Optimization Results}
\label{app:nova}

\begin{table}[h]
\centering
\caption{Held-out test scores on Amazon Nova Lite (mean over 3 repeats; 100 test questions; LLM-judge scale 0--100; same task acronyms as Table~\ref{tab:optimization}). 14 of 24 method$\times$task means fall below zero-shot; HelpSteer2 collapses from 6/6 (Haiku) to 1/6 here, illustrating \S\ref{sec:model_specificity}.}
\label{tab:nova}
\small
\begin{tabular}{@{}lcccc@{}}
\toprule
Method & FB & HS2 & WB & XSum \\
\midrule
Zero-Shot      & 80.4 & 70.7 & 64.6 & 73.5 \\
APE            & 81.1 & 69.4 & 64.4 & \textbf{73.9} \\
OPRO           & \textbf{81.9} & 70.0 & 64.2 & 73.5 \\
EvoPrompt      & 81.0 & 69.7 & 62.9 & 71.8 \\
PromptBreeder  & 80.2 & \textbf{72.8} & \textbf{65.6} & 72.9 \\
DSPy-style     & 81.0 & 69.1 & 60.2 & 73.3 \\
PROSE          & 80.4 & 70.0 & 64.6 & 72.8 \\
\bottomrule
\end{tabular}
\end{table}

\section{PROSE Method Details}
\label{app:prose}

PROSE (PRompt Optimization via Structured Evolution) decomposes each prompt into five semantic components (\texttt{role}, \texttt{task}, \texttt{constraints}, \texttt{examples}, \texttt{format}), enabling targeted modification of individual components while preserving others.

\paragraph{Seed generation.}
20 diverse candidates are generated using varied temperatures and prompting strategies, including a \emph{flat-then-decompose} approach.
The top 10 seeds (by training score) form the initial population.

\paragraph{Operators.}
Six operators with adaptive weights: targeted mutation (25\%), LLM crossover (20\%), random mutation (20\%), exploration (15\%), simplification (15\%), and random generation (5\%).
Weights shift toward operators whose offspring score higher (blend rate 0.3).

\paragraph{Selection.}
Candidates are ranked by a risk-adjusted fitness:
\begin{equation}
    \text{Fitness}(p) = 0.70 \cdot \bar{s}_p + 0.15 \cdot \widehat{\text{SR}}(p) + 0.15 \cdot \widehat{\text{DRO}}(p)
\end{equation}
where $\bar{s}_p$ is mean score, $\widehat{\text{SR}}$ is normalized Sharpe ratio, and $\widehat{\text{DRO}}$ penalizes worst-case failures.
Population size is 20 (elite 5), with early stopping after 4 generations without improvement (minimum 5 generations).

Despite this explicit risk-aware design, PROSE shows no measurable robustness advantage over simpler methods, consistent with our main finding that optimization gains are fragile and model-specific.

\end{document}